# Boosting in the presence of label noise

**Jakramate Bootkrajang**
School of Computer Science
University of Birmingham
Birmingham, B15 2TT
United Kingdom

**Ata Kabán**
School of Computer Science
University of Birmingham
Birmingham, B15 2TT
United Kingdom

## Abstract

Boosting is known to be sensitive to label noise. We studied two approaches to improve AdaBoost's robustness against labelling errors. One is to employ a label-noise robust classifier as a base learner, while the other is to modify the AdaBoost algorithm to be more robust. Empirical evaluation shows that a committee of robust classifiers, although converges faster than non label-noise aware AdaBoost, is still susceptible to label noise. However, pairing it with the new robust Boosting algorithm we propose here results in a more resilient algorithm under mislabelling.

## 1  Introduction

It is well known to practitioners that boosting is sensitive to label noise. The issue stems directly from the fundamental concept of boosting in that the effort is directed towards classifying the difficult samples. In fact, the complexity of the traditional boosting is very high, so much so that for a dataset with any configuration of its labels, it is possible to draw a decision boundary with zero training error. This seems to be a good approach to the classification problem if the difficult samples are not mislabelled samples in the first place. In reality however, there are no firm guarantees about the correctness of the class labels provided with the training set. In many applications, such as e.g. crowdsourcing data, and certain biomedical data, perfect training labels are almost impossible to obtain. A seemingly straightforward way to control boosting's complexity is by means of regularisation. However, regularisation alone might not be enough to solve this issue – as pointed out in Long & Servedio (2010), random misclassification defeats all boosters that optimise a convex objective. Yet, rather curiously, almost all of the existing *robust* boosters are still optimising a convex exponential loss. These boosters include the LogitBoost by Friedman et al. (1998) that optimises the binary log-loss, the Gentle-AdaBoost by Friedman et al. (1998) that is more stable because of a more conservative update step; the Modest-AdaBoost by Vezhnevets & Vezhnevets (2005) which penalises the ensemble when it makes a correct prediction on previously correctly predicted instances; the BB algorithm by Krieger et al. (2001) in which bagging is combined with boosting to average out the adverse effect of noisy labelled data. There is also a heuristic approach by Karmaker & Kwek (2006) where too difficult samples, i.e., those with very high weights, are removed from the training set according to a predefined threshold.

Motivated by the finding of Long and Servedio, Freund (2009) proposed a more robust boosting algorithm which optimises a non-convex potential function instead of the traditional exponential loss function. The general idea is to incorporate an early stopping as well as a mechanism to give up if the instance is to far away on the wrong side of the decision boundary. It shows promising results but unfortunately the boosting process becomes more complicated in that it also introduces a free parameter that has to be tuned. Freund suggests using cross-validation to tune the parameter however we can not rely on the cross-validation if our labels are noisy, unless we have a trusted validation set with correct labels.

Inspired by the work of Long and Servedio and Freund, we propose a different modification to AdaBoost for tackling label noise. We engineer our objective to be a combination of two complementary loss functions. Our new objective is somewhat related to those employed in the cost-sensitive boosting literature. However, in cost sensitive literature the cost associated with each instance is assumed to be given or known prior to the learning, and there is no label noise involved. By contrast, the primary task in robust boosting is to learn the mislabelling probabilities (which could be seen to be analogous to costs).

Recent developments on label-noise robust classifiers such as the robust Fisher Discriminant by Lawrence & Schölkopf (2001); Bouveyron & Girard (2009), the robust Logistic Regression by Bootkrajang & Kabán (2012); Raykar et al. (2010), the robust Gaussian Process by Hernández-Lobato et al. (2011) or the robust Nearest Neighbours by Barandela & Gasca (2000) suggest a new possibility to improve the existing booster without making any adjustment to the boosting algorithm by employing a robust classifier as a base learner. To the best of our knowledge, there are no attempts in the literature to pursue this direction and this is our starting point in this work.

To summarise, we investigate the solution to boosting in the presence of *random misclassification noise* at two different levels. At the lower level we study the robust committee where robust classifiers are combined and boosted using existing AdaBoost algorithm. At the higher level, we propose a new robust boosting algorithm that we call 'rBoost' where the objective function is a convex combination of two exponential losses. The coefficients of the combination represent uncertainty in the observed labels. The new boosting algorithm is closely related to AdaBoost and requires a relatively minor modification to the existing algorithm. Moreover our new objective is non-convex and exhibits robustness to labelling errors.

The paper is organised as follows. Section 2 reviews recent literature in label-noise robust classifiers and introduces the robust classifier that will be used throughout the paper. Section 3 presents the rBoost algorithm. Section 4 reports experimental results, and Section 6 draws conclusions of the study.

## 2 A robust base learner

In recent years many classifiers have been introduced to tackle the problem of learning in the presence of label noise. To date, there are a number of classifiers developed specifically for dealing with label noise: robust logistic regression, robust fisher discriminant, robust Gaussian Process or robust Nearest Neighbours. All of these can potentially be used as a base classifier, and it is then interesting to see how would such classifiers behave collectively in an ensemble. One way to construct a robust classifier is through a probabilistic latent variable model. Under the model, a robust classifier attempts to learn a posterior probability of the true labels via the likelihood of the observed labels.

We will deal with random label flipping noise, that is we assume the noise is independent of the specific features of individual data points, and flips the latent true label $y \in \{0,1\}$ from class $k$ to class $j$ into the observed label $\tilde{y} \in \{0,1\}$, with probability $\omega_{jk} := p(\tilde{y} = k|y = j)$. We define the likelihood of the observed label $\tilde{y}$ of a point $\mathbf{x}_i$ given the current parameter setting as the following:

$$\tilde{P}_i^k = p(\tilde{y} = k|\mathbf{x}_i, \theta, \{\omega_{jk}\}_{j,k=0}^1) = \sum_{j=0}^1 \omega_{jk} p(y = j|\mathbf{x}_i, \theta) \quad (1)$$

that is, a linear combination of the 'true' class posteriors. From this assumption the modified log-likelihood is given by

$$\mathcal{L}(\theta, \{\omega_{jk}\}_{j,k=0}^1) = \sum_{i=1}^n \sum_{k=0}^1 \mathbb{1}(\tilde{y}_i = k) \log(\tilde{P}_i^k) \quad (2)$$

where $\mathbb{1}(\cdot)$ is 1 if its argument is true and 0 otherwise, and $n$ is the number of training points. Note that any probabilistic classifier yielding class posterior probability will fit the framework and can be converted into a robust classifier using the technique shown.

For the sake of concreteness we will employ logistic regression with parameter $\theta = \boldsymbol{\beta}$ in our study. We call the model 'robust Logistic Regression' (rLR), in which the likelihood of $\tilde{y} = 1$ is defined as:

$$\tilde{P}_i^1 = \omega_{11}\sigma(\boldsymbol{\beta}^T\mathbf{x}_i) + \omega_{01}(1 - \sigma(\boldsymbol{\beta}^T\mathbf{x}_i)) \quad (3)$$

Here, $\boldsymbol{\beta}$ is the weight vector orthogonal to the decision boundary and it determines the orientation of the separating plane and $\sigma(a) = 1/(1+\exp(-a))$ is the sigmoid function. Learning the robust logistic regression model involves estimating $\boldsymbol{\beta}$ and well as $\omega_{jk}$. We follow the steps in Bootkrajang & Kabán (2012) where the conjugate gradient method is used to optimise $\boldsymbol{\beta}$. The gradient of the log-likelihood w.r.t the weight vector is, $\nabla_{\boldsymbol{\beta}}\mathcal{L}(\theta, \{\omega_{jk}\}_{j,k=0}^1) =$

$$\sum_{i=1}^n \left[ \left( \frac{\tilde{y}_i(\omega_{11} - \omega_{01})}{\tilde{P}_i^1} + \frac{(1-\tilde{y}_i)(\omega_{10} - \omega_{00})}{\tilde{P}_i^0} \right) \right.$$
$$\left. \times \sigma(\boldsymbol{\beta}^T\mathbf{x}_i)(1 - \sigma(\boldsymbol{\beta}^T\mathbf{x}_i)) \times \mathbf{x}_i \right] \quad (4)$$

The following multiplicative updates are then used to estimate $\omega_{jk}$:

$$\omega_{10} = \frac{g_{10}}{g_{10} + g_{11}}, \quad \omega_{11} = \frac{g_{11}}{g_{10} + g_{11}} \quad (5)$$

$$\omega_{00} = \frac{g_{00}}{g_{00} + g_{01}}, \quad \omega_{01} = \frac{g_{01}}{g_{00} + g_{01}} \quad (6)$$

where

$$g_{11} = \omega_{11} \sum_{i=1}^{n} \left( \frac{\tilde{y}_i \sigma(\boldsymbol{\beta}^T \mathbf{x}_i)}{\tilde{P}_i^1} \right)$$

$$g_{10} = \omega_{10} \sum_{i=1}^{n} \left( \frac{(1-\tilde{y}_i)\sigma(\boldsymbol{\beta}^T \mathbf{x}_i)}{\tilde{P}_i^0} \right)$$

$$g_{01} = \omega_{01} \sum_{i=1}^{n} \left( \frac{\tilde{y}_i(1-\sigma(\boldsymbol{\beta}^T \mathbf{x}_i))}{\tilde{P}_i^1} \right)$$

$$g_{00} = \omega_{00} \sum_{i=1}^{n} \left( \frac{(1-\tilde{y}_i)(1-\sigma(\boldsymbol{\beta}^T \mathbf{x}_i))}{\tilde{P}_i^0} \right) \quad (7)$$

## 3 The Robust Boosting

Suppose we have a training set with corrupted labels $D = \{(\mathbf{x}_i, \tilde{y}_i)\}_{i=1}^n$, where $\mathbf{x}_i \in \Re^m$ and $\tilde{y}_i \in \{+1, -1\}$. Let a base hypothesis be a decision function $h : \mathbf{x} \to \tilde{y}$. Under the boosting framework, a final hypothesis is a linear combination of the base hypotheses and it takes the following additive form:

$$H(x) = \sum_{t=1}^{T} \alpha_t h_t(\mathbf{x}) \quad (8)$$

In boosting, the 0/1 misclassification loss incurred by the final hypothesis is measured by the exponential loss:

$$\sum_{i=1}^{n} \mathbb{1}(\tilde{y}_i = 1)e^{-H(\mathbf{x}_i)} + \mathbb{1}(\tilde{y}_i = -1)e^{H(\mathbf{x}_i)} \quad (9)$$

This forms a boosting objective that has to be optimised. However, in the situation where labels are contaminated the loss in eq.(9) is not ideal, for obvious reasons. Instead, we form a new objective that explicitly takes into account uncertainties in labels:

$$\sum_{i=1}^{n} \mathbb{1}(\tilde{y}_i = 1)\left\{\gamma_{00}e^{-H(\mathbf{x}_i)} + \gamma_{01}e^{H(\mathbf{x}_i)}\right\}$$
$$+ \mathbb{1}(\tilde{y}_i = -1)\left\{\gamma_{11}e^{H(\mathbf{x}_i)} + \gamma_{10}e^{-H(\mathbf{x}_i)}\right\} \quad (10)$$

Here, $\gamma_{jk} = p(\tilde{y} = k | y = j)$ are probabilistic factors representing uncertainties in labels. Intuitively, the loss is weighed up or down depending on the gamma parameters $\gamma_{jk}$. For example, $\gamma_{01} = 0.3$ and $\gamma_{10} = 0$ indicates the situation where labels in the negative class (or class 0) are all correct – because no flipping from positive to negative occurred – but labels in the positive class are contaminated. Accordingly, the new loss accounts for this by adjusting the loss for the positive class (class 1) to: $0.7 * e^{-H} + 0.3 * e^H$. This is a hyperbolic cosine with the two tails adjusted and it represents the modified loss associated with the positive class. The shapes of such modified loss functions are depicted in Figure 1. From the figure we see that the classification that is 'too correct' will be penalised, hence reducing the overfitting problem. Meanwhile the loss of the negative class (class 0), which is $e^H + 0 * e^{-H} = e^H$, reduces to traditional boosting. It may be interesting to note that a similar shape of the loss can also be obtained by truncating the Taylor expansion of the exponential function to some finite degree. This could also be used to implement the same idea, although it would not have the transparent formulation given above.

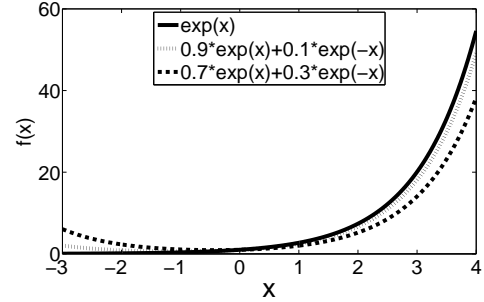

Figure 1: Various setups of the $\Gamma$ and their associated loss shape.

### 3.1 Adding a new base learner $h_t(\cdot)$

Consider the case $\tilde{y} = 1$, and define $d_{00} = e^{-H(\mathbf{x})}$, $d_{01} = e^{H(\mathbf{x})}$. Likewise, when $\tilde{y} = -1$ define $d_{11} = e^{H(\mathbf{x})}, d_{10} = e^{-H(\mathbf{x})}$ to be an unnormalised distribution of the data $(\mathbf{x}, \tilde{y})$. It can be shown that at the iteration $t$ of boosting, minimising the loss in eq.(10) w.r.t the new $h_t(\mathbf{x})$ is equivalent to minimising the following (the derivation details are given in the Appendix):

$$\arg\min_{h,\alpha} \, 2\sinh(\alpha) \sum_{i=1}^{n} \left\{ w_i \mathbb{1}(h(\mathbf{x}_i) \neq \tilde{y}_i) \right\}$$
$$+ e^{-\alpha} \sum_{i=1}^{n} \left\{ \mathbb{1}(\tilde{y}_i = 1)w_{00} + \mathbb{1}(\tilde{y}_i = -1)w_{11} \right\}$$
$$+ e^{\alpha} \sum_{i=1}^{n} \left\{ \mathbb{1}(\tilde{y}_i = 1)w_{01} + \mathbb{1}(\tilde{y}_i = -1)w_{10} \right\}$$
$$(11)$$

where

$$w_i = \begin{cases} (w_{00} - w_{01}), & \text{if } \tilde{y}_i = +1. \\ (w_{11} - w_{10}), & \text{if } \tilde{y}_i = -1. \end{cases} \quad (12)$$

and

$$w_{jk} = \gamma_{jk} \cdot d_{jk} \quad (13)$$

From this, it is immediate to see that in order to minimise the loss we have to seek for $h_t(\mathbf{x})$ that minimises the misclassification error $\epsilon_t = \sum_{i=1}^{n} w_i \mathbb{1}(\tilde{y}_i \neq h(\mathbf{x}_i))$.

This step is identical to the traditional AdaBoost except that the misclassification error of the current classifier is measured against different weighting factors which take into account the uncertainty of the observed noisy label as indicated by $\gamma_{jk}$. Note that the expression is fully compatible with the traditional AdaBoost such that the rBoost reduces to the original AdaBoost when $\gamma_{01} = 0$ and $\gamma_{10} = 0$. We emphasise that the weights in the rBoost need not be normalised. In fact in the original AdaBoost the normalisation simply facilitates the algebra in deriving a closed-form update for $\alpha_t$.

### 3.2 Updating $\alpha_t$

Now in our case, to get the update for $\alpha_t$ we take derivative of eq.(11) w.r.t $\alpha_t$, equate it to zero:

$$2\cosh(\alpha) \sum_{i=1}^{n} \left\{ w_i \mathbb{1}(h(\mathbf{x}_i) \neq \tilde{y}_i) \right\}$$
$$- e^{-\alpha} \sum_{i=1}^{n} \left\{ \mathbb{1}(\tilde{y}_i = 1)w_{00} + \mathbb{1}(\tilde{y}_i = -1)w_{11} \right\}$$
$$+ e^{\alpha} \sum_{i=1}^{n} \left\{ \mathbb{1}(\tilde{y}_i = 1)w_{01} + \mathbb{1}(\tilde{y}_i = -1)w_{10} \right\} = 0$$
(14)

Now, this equation cannot be solved in closed form. We resort to numerical optimisation to solve for the $\alpha_t$. Note the term which gets multiplied by $2\cosh(\alpha)$ is nothing but our error $\epsilon_t$ defined earlier.

### 3.3 Updating the sample weights

Next, to derive the update for the weight vectors, recall that we define $w_{jk} = \gamma_{jk} e^{-\tilde{y}_i H(\mathbf{x}_i)}$. It follows, for example, that the update for $w_{00}$ can be written as:

$$\begin{aligned}
w_{00}^{t+1} &= \gamma_{00} e^{-\tilde{y}_i(H + \alpha h)} \\
&= \gamma_{00} e^{-\tilde{y}_i H} \cdot e^{-\tilde{y}_i \alpha h} \\
&= \gamma_{00} d_{00}^t \cdot e^{\alpha(2\mathbb{1}(h(\mathbf{x}_i) \neq \tilde{y}_i) - 1)} \\
&= \gamma_{00} d_{00}^t \cdot e^{2\alpha \mathbb{1}(h(\mathbf{x}_i) \neq 1)} \cdot e^{-\alpha} \\
&\propto \gamma_{00} d_{00}^t \cdot e^{2\alpha \mathbb{1}(h(\mathbf{x}_i) \neq 1)}
\end{aligned} \quad (15)$$

where we used the rewriting: $-\tilde{y}h = 2\mathbb{1}(h(\mathbf{x}) \neq \tilde{y}) - 1$; and since $e^{-\alpha}$ are shared among all $w_{jk}$ it does not affect the optimisation. Similarly for the rest of the weight vectors we get:

$$w_{01}^{t+1} = \gamma_{01} d_{01}^t \cdot e^{2\alpha \mathbb{1}(h(\mathbf{x}_i) \neq -1)} \quad (16)$$
$$w_{11}^{t+1} = \gamma_{11} d_{11}^t \cdot e^{2\alpha \mathbb{1}(h(\mathbf{x}_i) \neq -1)} \quad (17)$$
$$w_{01}^{t+1} = \gamma_{10} d_{10}^t \cdot e^{2\alpha \mathbb{1}(h(\mathbf{x}_i) \neq 1)} \quad (18)$$

One way to implement this is to keep the distribution $d_{jk}$ separately and multiply it by $\gamma_{jk}$ to get a new $w_{jk}$ in each iteration.

### 3.4 Updating $\gamma_{jk}$

Finally, as mentioned earlier, we would also like to estimate the label flipping coefficients, $\gamma_{jk}$. We could take derivative of the loss incurred by the current ensemble, eq.(10), w.r.t each gamma and try to solve this direcly. This did not yield satisfactory results in our experience, most likely because the loss lacks probabilistic semantics. The workaround is to convert the output of boosting i.e. $H$ into a probability. There are three popular approaches to do that: 1) Logistic calibration $p(y = 1|x, H) = 1/(1 + \exp(-H))$ Friedman et al. (1998), 2) Platt's calibration $p(y = 1|x, H) = 1/(1 + \exp(AH + B))$ where $A$ and $B$ need to be learnt Platt (1999), and 3) Isotronic regression Robertson (1988). Niculescu-Mizil & Caruana (2005) empirically shows that Platt's technique and Isotronic regression are superior to a simple logistic transform. In addition, Platt's method has a slight advantage over IsoReg on small sample size. Hence, in this study, we will employ Platt's method to get calibrated posterior probabilities.

By converting $H$ to $p(y = 1|x, H)$, we can estimate the gamma from the following binomial log-loss, or cross-entropy. Using the notation $P(x) = p(y = 1|x, H)$ and $\bar{P}(x) = 1 - P(x)$, this is:

$$-\sum_{i=1}^{n} \mathbb{1}(\tilde{y}_i = 1) \log \left\{ \gamma_{11} P(\mathbf{x}_i) + \gamma_{01} \bar{P}(\mathbf{x}_i) \right\}$$
$$+ \mathbb{1}(\tilde{y}_i = -1) \log \left\{ \gamma_{00} \bar{P}(\mathbf{x}_i) + \gamma_{10} P(\mathbf{x}_i) \right\} \quad (19)$$

Following the Lagrangian method which imposes $\gamma_{00} + \gamma_{01} = 1$ and $\gamma_{11} + \gamma_{10} = 1$, similarly to the technique in the latent variable model (outlined in Section 2), the multiplicative updates for $\gamma_{jk}$ are found to be:

$$\gamma_{10} = \frac{g_{10}}{g_{10} + g_{11}}, \quad \gamma_{11} = \frac{g_{11}}{g_{10} + g_{11}} \quad (20)$$
$$\gamma_{00} = \frac{g_{00}}{g_{00} + g_{01}}, \quad \gamma_{01} = \frac{g_{01}}{g_{00} + g_{01}} \quad (21)$$

where

$$g_{11} = \gamma_{11} \sum_{i=1}^{n} \left( \frac{\mathbb{1}(\tilde{y}_i = 1) P_i}{\gamma_{11} P_i + \gamma_{01} \bar{P}_i} \right)$$
$$g_{10} = \gamma_{10} \sum_{i=1}^{n} \left( \frac{\mathbb{1}(\tilde{y}_i = -1) P_i}{\gamma_{10} P_i + \gamma_{00} \bar{P}_i} \right)$$
$$g_{01} = \gamma_{01} \sum_{i=1}^{n} \left( \frac{\mathbb{1}(\tilde{y}_i = 1) \bar{P}_i}{\gamma_{11} P_i + \gamma_{01} \bar{P}_i} \right)$$
$$g_{00} = \gamma_{00} \sum_{i=1}^{n} \left( \frac{\mathbb{1}(\tilde{y}_i = -1) \bar{P}_i}{\gamma_{10} P_i + \gamma_{00} \bar{P}_i} \right) \quad (22)$$

Our method is summarised in Algorithm 1.

As mentioned in the introduction, our rBoost algorithm has some analogies with cost-sensitive boosting Fan & Stolfo (1999); Masnadi-Shirazi & Vasconcelos (2011). One major difference is that the weighting factors in our case are outside of the exponential, whereas they are inside the exp in the mentioned works. In Fan & Stolfo (1999) the author did briefly discuss the possibility of having the weighting factors outside of the exponential, however their update of the weight is different from ours. Besides, the goal of cost-sensitive methods is different from ours. In cost-sensitive framework the cost is assumed to be known or given by the expert, and there is no implication of labelling errors.

**Algorithm 1** rBoost
  **Input:** data $\{\mathbf{x}, \tilde{y}\}^n$, boosting round $T$
  Initialize $w_{jk} = \gamma_{jk}$
  **for** $t = 1$ **to** $T$ **do**
    (1) $h_t = \arg\max_{\boldsymbol{\beta}}$ eq.(2) weighted by $w_i$.
    (2) Calculate the error w.r.t $w_i$ defined in eq.(12)
       $\epsilon_t = \sum_{i=1}^n w_i \mathbb{1}(\tilde{y}_i \neq h_t(\mathbf{x}_i))$
    (3) Optimise $\alpha_t$ numerically using the gradient in eq.(14)
    (4) Update $w_{jk}$ according to eq.(15)–(18).
    (5) Calculate $p(y = 1|\mathbf{x}, H)$ using Platt's method.
    (6) Update $\gamma_{jk}$ using eq.(20)-(21).
  **end for**
  Output the final classifier $sign(\sum_{t=1}^T \alpha_t h_t)$.

## 4 Empirical Evaluation

This section will investigate the performance of our robust boosting methods in practice. In addition, our new rBoost algorithm will be compared to the standard AdaBoost, GentleAdaBoost and ModestAdaBoost.

### 4.1 Methodology

We will study 4 configurations of base-learner and booster pairs: 1) LR + AdaBoost, 2) rLR + AdaBoost, 3) LR + rBoost and 4) rLR + rBoost. These four combinations will shed light on whether 1) a robust committee is robust against label errors?, 2) the new rBoost can counteract the bad effects of label noise? and finally 3) What can we get from pairing them together? We set our baseline to be the GentleAdaBoost and ModestAdaBoost where the base learner is a decision tree with maximum node splits of 2. For LR to serve as a weak learner, we employ random subsampling to create diversity in the ensemble. Further, we create two types of training sets by artificially injecting symmetric and asymmetric label noise at rate 10%, as well as at rate 30% into the training data. We train on the corrupted training set and validate the performance of the ensemble on a clean test set. We report the average and standard deviation of the misclassification rates from 10 independent random repetitions of 150 rounds of boosting each.

Table 1: Characteristic of the dataset used.

| Data set | # of pos. samp. | # of neg. samp. | dim. |
|---|---|---|---|
| Banana | 2375(45%) | 2924(55%) | 2 |
| Diabetes | 268(35%) | 500(65%) | 8 |
| Heart | 120(44%) | 150(56%) | 13 |
| Image | 1188(57%) | 898(43%) | 18 |
| Titanic | 14(58%) | 10(42%) | 3 |
| Twonorm | 3703(50%) | 3697(50%) | 20 |
| Waveform | 1647(33%) | 3353(67%) | 21 |

### 4.2 Datasets

We select seven UCI machine learning datasets Frank & Asuncion (2010) namely Banana, Diabetes, Heart, Image, Twonorm and Waveform to use to evaluate the proposed boosting combinations. We use 80% of the dataset for training and 20% for testing purpose. The characteristics of the datasets used are summarised in Table 1.

### 4.3 Results

We first investigate the behaviour of the robust classifiers as weak learners within the original AdaBoost algorithm. From the leftmost column of Tables 2-5, we see that when a robust classifier is used as a base learner the generalisation error of the ensemble is already lower in comparison to the original non-robust AdaBoost in 4 out of 7 datasets. The finding is consistent across all noise levels. It is very interesting to observe this because even though the base classifier is robust, it is still under the control of the original AdaBoost. Namely, the boosting will still guide the classifiers to focus on the more difficult parts of the dataset (which of course are likely to contain the points whose labels are wrong). Why is then this committee of robust classifiers more accurate? Lower error can come from two different sources: Either the robust committee is indeed robust against labelling errors, or it simply converges faster. To check this we run both configurations for more rounds to see the dynamics of the ensemble. Plotted in Figure 2 are the training and test errors of AdaBoost using the robust classifiers (rLR) as well as using the traditional classifiers (LR) on selected datasets. Superimposed for reference are ModestAdaBoost and rLR + rBoost.

It turns out that the robust committee converges much quicker than the non-robust committee. How-ever when boosted long enough we are starting to see that their classification performances become very similar. This answers our first research question. The robust classifiers as weak learners introduce what is understood to be a 'good diversity' in the ensemble, and drives the ensemble to convergence much quicker than the non-robust committee. Unfortunately however, the robustness of the base learner is not enough to withstand the effect of labelling errors.

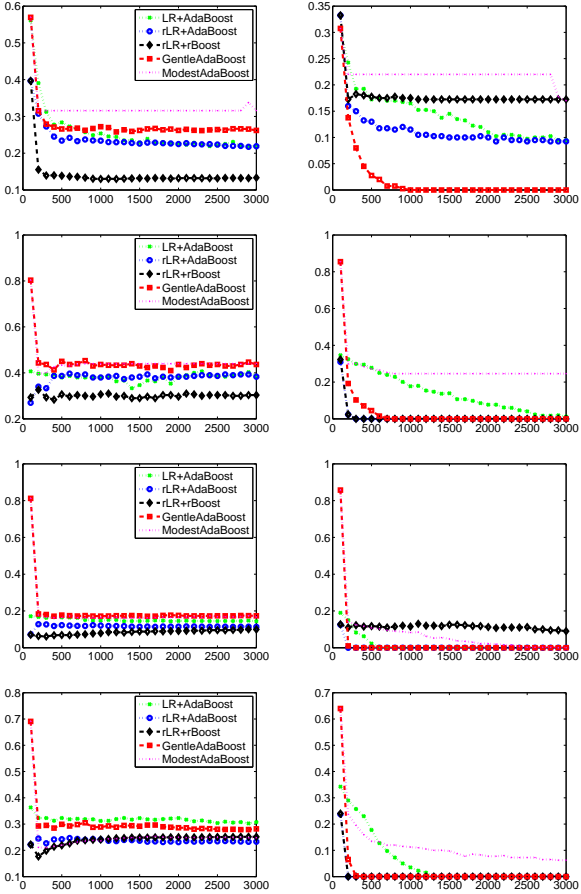

Figure 2: Test error(left) and traing error(right) for 'Banana','Diabetes','Twonorm','Waveform' datasets. The x-axis indicates boosting rounds while the y-axis shows classification errors.

Now we see that having rLR as a base learner alone is not enough to counteract the bad effect of mislabelling. We investigate further if we can pair rLR which has a fast convergence rate with our new rBoost algorithm.

Before proceeding, we need to establish that rBoost is superior to original AdaBoost when there is label noise. To this end, we consider two combinations: 1) AdaBoost+LR and 2)rBoost + LR in Tables 2-5. From the tables we see that rBoost+LR performs comparably to its non-robust booster counterpart when the noise rate is relatively low, and in the case of symmetric label case (i.e. the easy cases in terms of label noise). However when the noise is asymmetric and more severe (Table 5), rBoost significantly outperforms the original AdaBoost in all the cases tested. This answers our second research question. That is, rBoost significantly improves over the original AdaBoost in terms of classification performance especially in higher label contamination rate conditions and in asymmetric label noise conditions (i.e. the difficult cases).

Next, we equip our rBoost method with the robust base classifiers that enjoy fast convergence to obtain our final robust boosting algorithm. These results are shown in the fourth column of Table 5. The superior performance of this approach is most apparent, and we also give an illustrative example of the working of our rBoost on the 'Banana' dataset in Figure 3. We see that the original AdaBoost generated a patchy decision boundary as a result of label noise, while our rBoost returned a smoother and more appropriate decision boundary.

Further, we validate our approach for estimating the flip probabilities $\gamma_{jk}$ using the multiplicative updates given in eq.(20) and eq.(21). Disappointingly, we see that the results (5th and 6th column of Tables 2-5) are not as good as the ideal setting where the $\gamma_{jk}$ are fixed to the true value (rBoost-Fixed gamma). However, and more interestingly, we observe that the quality of the estimated gammas depends highly on the quality of the calibrated probability used in the update. Assuming that we have a trusted validation set that we can use to obtain a more accurate calibrated probability, we ask how well can we estimate the gammas? We hold out a small subset of the dataset, where all of the labels are clean. This will be our trusted validation set, and we took this set as tiny as 20 points only. We feed this small trusted dataset into the Platt's calibration method. We carried out this experiment on Banana, Image and Twonorm. The classification error from 10 repeated runs of our rBoost algorithm with the use of the trusted validation set as a source for calibrating the probability is 15.74±0.23% on 'Banana' at 30% asymmetric noise, compared to 23.83% without. This is taken from the sixth column of Table 5. On Image at 30% asymmetric noise the error is as low as 7.61±0.19% and on Twonorm it is 9.73±0.31%. Intriguingly, a tiny trusted set of 20 points is able to improve the situation even for the Image data, where the training set size is as large as 1300 (80% of total number of samples in Image). Thus we can conclude that the trusted validation set approach may be seen as a technique to effectively and efficiently incorporate extra knowledge about the labels into the rBoost algorithm. We should note, this differs from simply in-cluding the trusted samples into the training set, since the latter would simply make a slight reduction of the noise rate. Of course, the larger the trusted validation set for calibration, the better probability calibration we can expect, and consequently this should lead to more accurate estimates of the gammas ($\gamma_{jk}$), and hence to better classification performance.

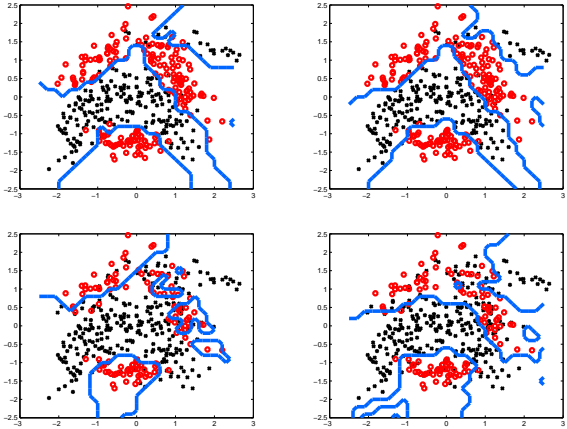

Figure 3: Comparison of the decision boundaries obtained from AdaBoost(left) and rBoost(right) in noise-free case(top) and 30% asymmetric noise case(bottom) on banana dataset.

## 5 Discussion

We have assumed throughout the study that the label noise is random and it occurs independently from the input sample. Worth mentioning that there exist other types of noise such as a non-random label noise, malicious or adverserial noise, which may require a different treatment. The random noise treated here is simpler and more generic as it does not require special knowledge about the noise process. By contrast, modelling a non-random noise requires us to encode domain expertise into the formulation, and as a consequence it will yield a model specific to the application. Interestingly, despite its simplicity, the random noise model finds its use even in the cases where the random assumption does not perfectly hold true, as it microarray anaylsis (Bootkrajang & Kabán (2013)). In the context of boosting, there is an attempt to tackle non-random noise in Takenouchi et al. (2008) where both binary and multi-class problems are investigated.

## 6 Conclusion

We presented a robust boosting algorithm based on the famous AdaBoost algorithm, which we called rBoost. The rBoost has some advantageous properties, namely its objective is non-convex, hence more robust, and it incorporates label noise parameters that can be estimated efficiently using the proposed multiplicative update rules. The new algorithm is also appealing since it requires a minor modification to the existing AdaBoost algorithm. We further demonstrated that the label noise parameters can be more accurately estimated by using a trusted validation set for Platt's calibration algorithm as a form of extra information. It shows good result close to the rBoost with label noise parameters fixed to the true values. In addition, we have empirically shown that simply employing a robust classifier as a base learner in the AdaBoost does not help alleviating the bad effect of label noise. However, rather interestingly, its effect is to speed up the boosting process. This could be advantageous in cases of low noise. An intriguing future direction are the theoretical analysis of our proposed rBoost and extensions to multi-class problems.

### Acknowledgements

We would like to thank the anonymous reviewers for their valuable comments and suggestions, and the School of Computer Science, University of Birmingham for providing computing facilities for this study. JB thanks the Royal Thai Government for financial support.

## Appendix

This section shows derivation details of the rBoost algorithm. The loss of the rBoost is defined as:

$$\mathcal{L}(H) = \sum_{i=1}^{n} \mathbb{1}(\tilde{y}_i = 1)\Big\{\gamma_{00}e^{-H(\mathbf{x}_i)} + \gamma_{01}e^{H(\mathbf{x}_i)}\Big\}$$
$$+ \mathbb{1}(\tilde{y}_i = -1)\Big\{\gamma_{11}e^{H(\mathbf{x}_i)} + \gamma_{10}e^{-H(\mathbf{x}_i)}\Big\} \quad (23)$$

Here, $\gamma_{jk}$ are again probabilistic factors representing uncertainty in labels. We write out the form of $H(\mathbf{x}_i)$ for the next round of AdaBoost. Minimising this loss of eq. (23) in a step-wise manner is then equivalent to minimising the following:

$$\arg\min_{h,\alpha}$$
$$\Big(\sum_{i=1}^{n} \mathbb{1}(\tilde{y}_i = 1)\Big\{\gamma_{00}e^{-(H(\mathbf{x}_i)+\alpha h(\mathbf{x}_i))} + \gamma_{01}e^{H(\mathbf{x}_i)+\alpha h(\mathbf{x}_i)}\Big\}$$
$$+ \mathbb{1}(\tilde{y}_i = -1)\Big\{\gamma_{11}e^{H(\mathbf{x}_i)+\alpha h(\mathbf{x}_i)} + \gamma_{10}e^{-(H(\mathbf{x}_i)+\alpha h(\mathbf{x}_i))}\Big\}\Big) \quad (24)$$

$$= \arg\min_{h,\alpha} \sum_{i=1}^{n} \Big\{\mathbb{1}(\tilde{y}_i = 1)\gamma_{00}e^{-H(\mathbf{x}_i)}e^{-\alpha h(\mathbf{x}_i)} \quad (25)$$
$$+ \mathbb{1}(\tilde{y}_i = 1)\gamma_{01}e^{H(\mathbf{x}_i)}e^{\alpha h(\mathbf{x}_i)} \quad (26)$$
$$+ \mathbb{1}(\tilde{y}_i = -1)\gamma_{11}e^{H(\mathbf{x}_i)}e^{\alpha h(\mathbf{x}_i)} \quad (27)$$
$$+ \mathbb{1}(\tilde{y}_i = -1)\gamma_{10}e^{-H(\mathbf{x}_i)}e^{-\alpha h(\mathbf{x}_i)}\Big\} \quad (28)$$

Table 2: Average classification errors and their standard deviations for AdaBoost and rBoost at 10% symmetric noise. Boldface font shows the result which is statistically significant as tested with Wilcoxon ranksum test at the 5% level.

| Dataset | AdaBoost | | rBoost-Fixed gamma | | rBoost | | Gentle Boost | Modest Boost |
|---|---|---|---|---|---|---|---|---|
| | LR | rLR | LR | rLR | LR | rLR | | |
| Banana | 18.53±1.0 | 13.13±1.1 | 17.53±1.8 | 12.94±0.9 | 17.44±1.8 | 12.96±0.9 | 16.09±1.6 | 21.87±3.4 |
| Diabetes | 24.00±2.7 | 25.80±2.3 | 24.10±1.7 | 25.63±1.5 | 23.87±1.9 | 25.20±2.4 | 27.40±2.0 | 24.33±1.9 |
| Heart | 21.20±4.4 | 21.60±3.1 | 22.40±3.9 | 20.30±3.5 | 22.10±4.8 | 20.90±4.4 | 23.60±3.1 | 22.40±3.5 |
| Image | 14.61±1.3 | 4.08±1.0 | 15.29±1.5 | 4.49±0.8 | 13.51±1.4 | 4.12±0.8 | 4.43±0.7 | 15.91±3.1 |
| Titanic | 22.76±1.3 | 22.32±1.1 | 22.97±1.4 | 22.37±1.1 | 22.76±1.2 | 22.37±1.4 | 22.30±1.7 | 23.28±1.4 |
| Twonorm | 5.78±0.8 | 4.30±0.8 | 5.75±0.7 | 4.42±0.9 | 5.72±1.0 | 4.41±0.7 | 9.65±1.0 | 7.21±0.5 |
| Waveform | 16.67±1.5 | **13.40±0.7** | 16.43±0.7 | 14.65±0.8 | 16.12±1.2 | **13.47±0.6** | 14.98±0.8 | 14.77±1.5 |
| All | 17.65±1.8 | **14.95±1.5** | 17.78±1.6 | **14.97±1.3** | 17.36±1.9 | **14.78±1.6** | 16.92±1.5 | 18.54±2.2 |

Table 3: Average classification errors and their standard deviations for AdaBoost and rBoost at 30% symmetric noise.

| Dataset | AdaBoost | | rBoost-Fixed gamma | | rBoost | | Gentle Boost | Modest Boost |
|---|---|---|---|---|---|---|---|---|
| | LR | rLR | LR | rLR | LR | rLR | | |
| Banana | 18.71±3.1 | 14.73±3.0 | 18.14±1.7 | 14.47±2.1 | 18.05±1.6 | 14.94±2.7 | 20.62±1.6 | 24.69±2.5 |
| Diabetes | 25.37±2.3 | 27.47±1.9 | 24.90±2.3 | 29.57±2.4 | 25.23±2.7 | 28.57±2.3 | 30.60±2.9 | 26.67±2.3 |
| Heart | 22.10±5.7 | 21.50±4.0 | 22.70±4.0 | 22.60±6.5 | 22.90±4.9 | 21.90±4.3 | 30.00±5.5 | 24.80±3.7 |
| Image | 14.67±1.4 | 6.94±1.0 | 15.23±1.0 | 6.67±1.0 | 14.30±0.9 | 6.52±0.9 | 7.51±1.0 | 20.10±4.4 |
| Titanic | 23.12±1.6 | 23.01±1.8 | 23.27±1.5 | 22.92±1.4 | 23.09±1.4 | 23.11±1.8 | 22.80±1.9 | 23.41±1.3 |
| Twonorm | 8.53±1.1 | 6.67±0.9 | 8.77±1.0 | 6.87±1.3 | 8.63±1.0 | 6.60±1.1 | 16.06±2.0 | 8.84±0.9 |
| Waveform | 21.02±2.1 | 16.88±1.8 | 20.80±2.4 | 18.16±2.0 | 20.41±2.2 | 16.96±1.6 | 20.26±2.2 | 15.57±0.9 |
| All | 19.07±2.5 | **16.74±2.1** | 19.11±1.9 | **17.32±2.4** | 18.94±2.1 | **16.94±2.1** | 21.12±2.4 | 20.58±2.3 |

Now consider each term in the sum.

$$(25) = \sum_{i|h(\mathbf{x}_i)=\tilde{y}_i} \mathbb{1}(\tilde{y}_i = 1)\gamma_{00} e^{-H(\mathbf{x}_i)} e^{-\alpha}$$
$$+ \sum_{i|h(\mathbf{x}_i)\neq \tilde{y}_i} \mathbb{1}(\tilde{y}_i = 1)\gamma_{00} e^{-H(\mathbf{x}_i)} e^{\alpha}$$
$$= \sum_{i=1}^{n} \left(1 - \mathbb{1}(h(\mathbf{x}_i) \neq \tilde{y}_i)\right) \mathbb{1}(\tilde{y}_i = 1)\gamma_{00} e^{-H(\mathbf{x}_i)} e^{-\alpha}$$
$$+ \sum_{i|h(\mathbf{x}_i)\neq \tilde{y}_i} \mathbb{1}(\tilde{y}_i = 1)\gamma_{00} e^{-H(\mathbf{x}_i)} e^{\alpha}$$
$$= \sum_{i=1}^{n} \mathbb{1}(\tilde{y}_i = 1)\gamma_{00} e^{-H(\mathbf{x}_i)} e^{-\alpha}$$
$$- \sum_{i=1}^{n} \mathbb{1}(\tilde{y}_i = 1)\mathbb{1}(h(\mathbf{x}_i) \neq \tilde{y}_i)\gamma_{00} e^{-H(\mathbf{x}_i)} e^{-\alpha}$$
$$+ \sum_{i=1}^{n} \mathbb{1}(\tilde{y}_i = 1)\mathbb{1}(h(\mathbf{x}_i) \neq \tilde{y}_i)\gamma_{00} e^{-H(\mathbf{x}_i)} e^{\alpha}$$
$$= \left(e^{\alpha} - e^{-\alpha}\right) \sum_{i=1}^{n} \mathbb{1}(\tilde{y}_i = 1)\mathbb{1}(h(\mathbf{x}_i) \neq \tilde{y}_i)\gamma_{00} e^{-H(\mathbf{x}_i)}$$
$$+ \sum_{i=1}^{n} \mathbb{1}(\tilde{y}_i = 1)\gamma_{00} e^{-H(\mathbf{x}_i)} e^{-\alpha} \quad (29)$$

Using similar substitution and grouping, we also have the following:

$$(26) = \left(e^{-\alpha} - e^{\alpha}\right) \sum_{i=1}^{n} \mathbb{1}(\tilde{y}_i = 1)\mathbb{1}(h(\mathbf{x}_i) \neq \tilde{y}_i)\gamma_{01} e^{H(\mathbf{x}_i)}$$
$$+ \sum_{i=1}^{n} \mathbb{1}(\tilde{y}_i = 1)\gamma_{01} e^{H(\mathbf{x}_i)} e^{\alpha} \quad (30)$$

$$(27) = \left(e^{\alpha} - e^{-\alpha}\right) \sum_{i=1}^{n} \mathbb{1}(\tilde{y}_i = -1)\mathbb{1}(h(\mathbf{x}_i) \neq \tilde{y}_i)\gamma_{11} e^{H(\mathbf{x}_i)}$$
$$+ \sum_{i=1}^{n} \mathbb{1}(\tilde{y}_i = -1)\gamma_{11} e^{H(\mathbf{x}_i)} e^{-\alpha} \quad (31)$$

$$(28) = \left(e^{-\alpha} - e^{\alpha}\right) \sum_{i=1}^{n} \mathbb{1}(\tilde{y}_i = -1)\mathbb{1}(h(\mathbf{x}_i) \neq \tilde{y}_i)\gamma_{10} e^{-H(\mathbf{x}_i)}$$
$$+ \sum_{i=1}^{n} \mathbb{1}(\tilde{y}_i = -1)\gamma_{10} e^{-H(\mathbf{x}_i)} e^{\alpha} \quad (32)$$

Summing all four expressions we have the objective:

$$\arg\min_{h,\alpha} (e^{\alpha} - e^{-\alpha}) \sum_{i=1}^{n} \Big\{ \mathbb{1}(\tilde{y}_i = 1)\gamma_{00} e^{-H(\mathbf{x}_i)} \mathbb{1}(h(\mathbf{x}_i) \neq \tilde{y}_i)$$
$$+ \mathbb{1}(\tilde{y}_i = -1)\gamma_{11} e^{H(\mathbf{x}_i)} \mathbb{1}(h(\mathbf{x}_i) \neq \tilde{y}_i) \Big\}$$
$$+ e^{-\alpha} \sum_{i=1}^{n} \Big\{ \mathbb{1}(\tilde{y}_i = 1)\gamma_{00} e^{-H(\mathbf{x}_i)} + \mathbb{1}(\tilde{y}_i = -1)\gamma_{11} e^{H(\mathbf{x}_i)} \Big\}$$
$$- (e^{\alpha} - e^{-\alpha}) \sum_{i=1}^{n} \Big\{ \mathbb{1}(\tilde{y}_i = 1)\gamma_{01} e^{H(\mathbf{x}_i)} \mathbb{1}(h(\mathbf{x}_i) \neq \tilde{y}_i)$$
$$+ \mathbb{1}(\tilde{y}_i = -1)\gamma_{10} e^{-H(\mathbf{x}_i)} \mathbb{1}(h(\mathbf{x}_i) \neq \tilde{y}_i) \Big\}$$
$$+ e^{\alpha} \sum_{i=1}^{n} \Big\{ \mathbb{1}(\tilde{y}_i = 1)\gamma_{01} e^{H(\mathbf{x}_i)} + \mathbb{1}(\tilde{y}_i = -1)\gamma_{10} e^{-H(\mathbf{x}_i)} \Big\}$$
$$\quad (33)$$

Table 4: Average classification errors and their standard deviations for AdaBoost and rBoost at 10% asymmetric noise.

| Dataset | AdaBoost | | rBoost-Fixed gamma | | rBoost | | Gentle Boost | Modest Boost |
|---|---|---|---|---|---|---|---|---|
| | LR | rLR | LR | rLR | LR | rLR | | |
| Banana | 17.85±2.7 | 13.54±1.3 | 16.78±1.7 | 12.55±1.1 | 17.81±2.4 | 13.65±1.2 | 16.81±1.5 | 22.27±2.8 |
| Diabetes | 24.70±1.6 | 25.67±1.6 | 24.27±1.7 | 25.57±1.5 | 24.23±1.6 | 25.97±1.7 | 27.80±2.1 | 24.93±1.6 |
| Heart | 21.70±4.4 | 21.50±3.9 | 21.60±3.7 | 22.20±2.9 | 21.30±3.7 | 21.10±2.7 | 26.10±3.2 | 21.70±4.2 |
| Image | 15.88±1.0 | 4.52±1.0 | 15.20±1.6 | 4.06±0.9 | 15.15±1.5 | 4.40±1.2 | 4.45±0.7 | 24.12±1.9 |
| Titanic | 22.87±1.3 | 23.06±1.3 | 22.65±1.3 | 22.23±1.1 | 23.58±1.6 | 22.44±1.0 | 22.83±1.5 | 23.63±1.5 |
| Twonorm | 7.02±1.6 | 5.21±1.1 | 6.16±1.4 | 4.51±0.8 | 6.56±1.4 | 5.34±1.1 | 9.19±1.0 | 7.71±1.2 |
| Waveform | 18.10±1.3 | 14.48±1.1 | 17.83±1.2 | 16.23±1.5 | 17.71±1.2 | 14.54±1.4 | 16.52±1.0 | 14.19±0.7 |
| All | 18.30±1.9 | **15.42±1.6** | 17.78±1.7 | **15.33±1.4** | 18.04±1.9 | **15.34±1.5** | 17.67±1.6 | 19.79±1.9 |

Table 5: Average classification errors and their standard deviations for AdaBoost and rBoost at 30% asymmetric noise.

| Dataset | AdaBoost | | rBoost-Fixed gamma | | rBoost | | Gentle Boost | Modest Boost |
|---|---|---|---|---|---|---|---|---|
| | LR | rLR | LR | rLR | LR | rLR | | |
| Banana | 31.45±5.2 | 23.53±4.7 | 27.31±4.5 | **14.27±1.0** | 32.39±3.9 | 23.83±4.1 | 25.38±2.7 | 33.04±6.8 |
| Diabetes | 32.20±2.1 | 33.43±3.9 | 29.47±3.0 | **30.20±2.6** | 32.80±3.3 | 33.27±3.1 | 38.37±3.6 | 32.07±3.5 |
| Heart | 27.60±5.5 | 27.30±6.8 | 23.00±4.3 | **24.30±3.8** | 28.20±6.3 | 28.00±6.9 | 32.00±7.2 | 29.60±11.7 |
| Image | 22.48±1.6 | 10.70±0.9 | 16.96±1.8 | **5.47±1.0** | 20.53±1.6 | 9.82±1.5 | 11.94±1.1 | 26.44±1.3 |
| Titanic | 32.60±8.4 | 31.21±8.7 | 23.88±1.8 | **22.14±1.5** | 33.17±9.3 | 30.73±8.7 | 32.94±9.0 | 33.49±13.7 |
| Twonorm | 16.02±2.4 | 12.07±2.0 | 8.89±1.5 | **6.51±1.3** | 14.68±2.3 | 12.19±2.0 | 17.85±1.8 | 16.62±3.1 |
| Waveform | 28.83±2.8 | 23.43±2.5 | 24.27±2.1 | **19.95±1.6** | 28.39±3.1 | 23.02±2.5 | 27.31±2.5 | 21.10±2.2 |
| All | 27.31±3.9 | 23.09±4.2 | 21.96±2.7 | **17.54±1.8** | 27.16±4.2 | 22.97±4.1 | 26.54±3.9 | 27.47±6.0 |

$$= \arg\min_{h,\alpha} 2\sinh(\alpha) \times$$

$$\sum_{i=1}^{n} \left\{ \mathbb{1}(\tilde{y}_i = 1)\mathbb{1}(h(\mathbf{x}_i) \neq \tilde{y}_i)[\gamma_{00} e^{-H(\mathbf{x}_i)} - \gamma_{01} e^{H(\mathbf{x}_i)}] \right\}$$

$$+ 2\sinh(\alpha) \times$$

$$\sum_{i=1}^{n} \left\{ \mathbb{1}(\tilde{y}_i = -1)\mathbb{1}(h(\mathbf{x}_i) \neq \tilde{y}_i)[\gamma_{11} e^{H(\mathbf{x}_i)} - \gamma_{10} e^{-H(\mathbf{x}_i)}] \right\}$$

$$+ e^{-\alpha} \sum_{i=1}^{n} \left\{ \mathbb{1}(\tilde{y}_i = 1)\gamma_{00} e^{-H(\mathbf{x}_i)} + \mathbb{1}(\tilde{y}_i = -1)\gamma_{11} e^{H(\mathbf{x}_i)} \right\}$$

$$+ e^{\alpha} \sum_{i=1}^{n} \left\{ \mathbb{1}(\tilde{y}_i = 1)\gamma_{01} e^{H(\mathbf{x}_i)} + \mathbb{1}(\tilde{y}_i = -1)\gamma_{10} e^{-H(\mathbf{x}_i)} \right\}$$

$$(34)$$

Define $w_{00} = \gamma_{00} e^{-H(\mathbf{x}_i)}$, $w_{01} = \gamma_{01} e^{H(\mathbf{x}_i)}$, $w_{11} = \gamma_{11} e^{H(\mathbf{x}_i)}$ and $w_{10} = \gamma_{10} e^{-H(\mathbf{x}_i)}$, we simplify the objective into.

$$\arg\min_{h,\alpha} 2\sinh(\alpha) \sum_{i=1}^{n} \left\{ w_i \mathbb{1}(h(\mathbf{x}_i) \neq \tilde{y}_i) \right\}$$

$$+ e^{-\alpha} \sum_{i=1}^{n} \left\{ \mathbb{1}(\tilde{y}_i = 1)w_{00} + \mathbb{1}(\tilde{y}_i = -1)w_{11} \right\}$$

$$+ e^{\alpha} \sum_{i=1}^{n} \left\{ \mathbb{1}(\tilde{y}_i = 1)w_{01} + \mathbb{1}(\tilde{y}_i = -1)w_{10} \right\} \quad (35)$$

where

$$w_i = \begin{cases} (w_{00} - w_{01}), & \text{if } \tilde{y}_i = +1. \\ (w_{11} - w_{10}), & \text{if } \tilde{y}_i = -1. \end{cases} \quad (36)$$